\documentclass{article}
\usepackage{spconf,amsmath,graphicx}
\usepackage{booktabs}

\usepackage{color}

\title{GENERAL AUDIO TAGGING WITH ENSEMBLING CONVOLUTIONAL NEURAL NETWORKS AND STATISTICAL FEATURES}

\name{Kele Xu$^{1,2}$,
	Boqing Zhu$^{1,2}$,
	Qiuqiang Kong$^{3}$,
	Haibo Mi$^{1,2}$,
	Bo Ding$^{1,2}$,
	Dezhi Wang$^{4}$\sthanks{Corresponding author.},
	Huaimin Wang$^{1,2}$
}

\address{$^1$ National Key Laboratory of Parallel and Distributed Processing, Changsha, China,\\
	$^2$ College of Computer, National University of Defense Technology, Changsha, China,\\
	$^3$ Centre for Vision, Speech and Signal Processing (CVSSP), University of Surrey, UK\\
	$^4$ College of Meteorology and Oceanography, National Univ. of Defense Tech, Changsha, China \\
	kelele.xu@gmail.com, wang\_dezhi@hotmail.com
}
\begin{document}
%\ninept
%
\maketitle
\begin{abstract}

Audio tagging aims to infer descriptive labels from audio clips.
Audio tagging is challenging due to the limited size of data and noisy labels.
In this paper, we describe our solution for the DCASE 2018 Task 2 general audio tagging challenge.  
The contributions of our solution include: We investigated a variety of convolutional neural network architectures to solve the audio tagging task.
Statistical features are applied to capture statistical patterns of audio features to improve the classification performance. Ensemble learning is applied to ensemble the outputs from the deep classifiers to utilize complementary information. a sample re-weight strategy is employed for ensemble training to address the noisy label problem. 
Our system achieves a mean average precision (mAP@3) of 0.958, outperforming the baseline system of 0.704. Our system ranked the 1st and 4th out of 558 submissions in the public and private leaderboard of DCASE 2018 Task 2 challenge. 
Our codes are available at https://github.com/Cocoxili/DCASE2018Task2/.

\end{abstract}
\section{Introduction}
\label{sec:intro}

Audio tagging task is a task to predict the presence or absence of certain acoustic events in an audio recording, and it has drawn lots of attention during the last several years. Audio tagging has widely applications, such as surveillance, monitoring, and health care \cite{Fonseca2018_DCASE}.
Historically, audio tagging has been addressed with different handcrafted features and shallow-architecture classifiers including Gaussian mixture models (GMMs) \cite{mesaros2016tut} and non-negative matrix factorizations (NMFs) \cite{mesaros2017dcase}. 
Recently, deep learning approaches such as convolutional neural networks (CNNs) have achieved state-of-the-art performance for the audio tagging task \cite{hershey2017CNN,xu2017unsupervised}.

DCASE 2018 Task 2 launched a competition for the general audio tagging task \cite{Fonseca2018_DCASE} to attract research interests for the audio tagging problem. However, due to the limited size of data and noisy labels \cite{Fonseca2018_DCASE}, general audio tagging remains as a challenge and falls short of accuracy and robustness. The current general audio tagging systems are confronted with several challenges: (1) There are a large amount of event classes in \cite{Fonseca2018_DCASE} compared with previous audio classification tasks \cite{mesaros2017dcase, foster2015chime}. (2) The imbalance problem could make the model emphasize more on the classes with more training samples and difficult to learn from the classes with less samples. (3) The data quality varies from class to class. For example, some audio clips are manually verified in \cite{Fonseca2018_DCASE} but others are not. Designing supervised deep learning algorithms that can learn from data sets with noisy labels is an important problem, especially, when the data set is small.

In this paper, we aim to build scalable ensemble approach with taking the noisy label into account. The proposed method achieves a state-of-the-art performance on the DCASE 2018 Task 2 dataset. The contributions of the paper are summarized as below:
(1) A quantitative comparison is investigated using different convolutional neural network (CNN) architectures inspired from computer vision. These CNN architectures are further deployed for the ensemble learning.
(2) We propose to employ statistical features including the skewness and kurtosis of frame-wise MFCC to improve the performance.
(3) A scalable ensemble approach is used to utilize the complementary information of different deep architectures and handcrafted features.
(4) A samples re-weight strategy is proposed for the ensemble learning to solve the noisy label problem in the dataset.

The paper is organized as follows: Section 2 describes the proposed CNNs, statistical features, ensemble learning and sample re-weight methods. Section 3 shows experimental results. Section 4 concludes and forecasts future work.

\section{METHODOLOGY}
\subsection{Convolutional neural networks}
CNNs have been successfully applied to many computer vision tasks \cite{simonyan2014very, szegedy2016rethinking, he2016deep}. Though there are many works using CNN for audio tagging \cite{mesaros2017dcase}, there is few work investigating a quantitative comparison of different CNNs on the audio tagging task.
In this paper, we investigated 7 effective CNN architectures from computer vision on the tagging task including VGG \cite{simonyan2014very}, Inception \cite{szegedy2016rethinking}, ResNet \cite{he2016deep}, DenseNet \cite{huang2017densely}, ResNeXt \cite{xie2017aggregated}, SE-ResNeXt \cite{hu2017squeeze} and DPN \cite{chen2017dual}. Our aim is to investigate these CNNs on the audio tagging task. 

VGGNet \cite{simonyan2014very} consists of 3$\times$3 convolutional layers stacked on top of each other to increase the depth of a CNN. 
Inception \cite{szegedy2016rethinking} applies different size of convolution filter within the blocks of a network, which can act as a ``multi-level feature extractor''.
ResNet \cite{he2016deep} introduces residual models to alleviate gradient vanishing problem to train very deep CNNs.
DenseNet \cite{huang2017densely} consists of many dense blocks, which are connected to a transition layer to re-utilize the previous features.
ResNeXt \cite{xie2017aggregated} is an improvement of ResNet. It is constructed by repeating a building block that aggregates a set of transformations with the same topology.
By introducing the Squeeze-and-Excitation (SE) block \cite{hu2017squeeze}, networks could improve the representational power by explicitly modeling the interdependencies between the channels of its convolutional features. The SE block can be deployed on the ResNeXt, which is denoted as SE-ResNeXt in this paper.
DPN inherits the benefits from ResNet and DenseNet. It shares common features while maintaining the flexibility to explore new features through dual path architectures.

It is worthwhile to notice that two different ways to train a deep model including: using ImageNet-based pre-trained model to initialize the weights and finetune the model or randomly initialize the weights and train the model from scratch.

\subsection{Statistical features}
Some statistical patterns of the audio representation cannot be easily learned by the deep models. For example, the higher-order statistics including the skewness and kurtosis. We show that these handcrafted statistical features can provide complementary information, which can be used to improve the classification performance \cite{fonseca2018simple}. 

In our experiments, as the audio pre-processing method described above, all audio samples are divided into 1.5-second audio clips.
We employ the statistical features on raw audio signal and MFCC features. The statistical features including the mean, variance, variance of the derivative, skewness and kurtosis. The definitions of skewness and kurtosis are given as follows.

\begin{equation}
Skewness=E\left\{(\frac{X-\mu}{\sigma})^3 \right\},
\end{equation}

\begin{equation}
Kurtosis=E\left\{(\frac{X-\mu}{\sigma})^4 \right\},
\end{equation}
where $X$ is the vector (for example, $X$ can be the raw signal or the MFCC of an audio segment which is random selected from the audio clip). $\mu$ is the mean, $\sigma$ is the standard deviation, and $E$ is the expectation operator. The statistical features are clip-wise, which suggests that the statistical analysis is conducted for each clip.

\begin{figure}[h]
	\centering
	\includegraphics[width=0.45\textwidth]{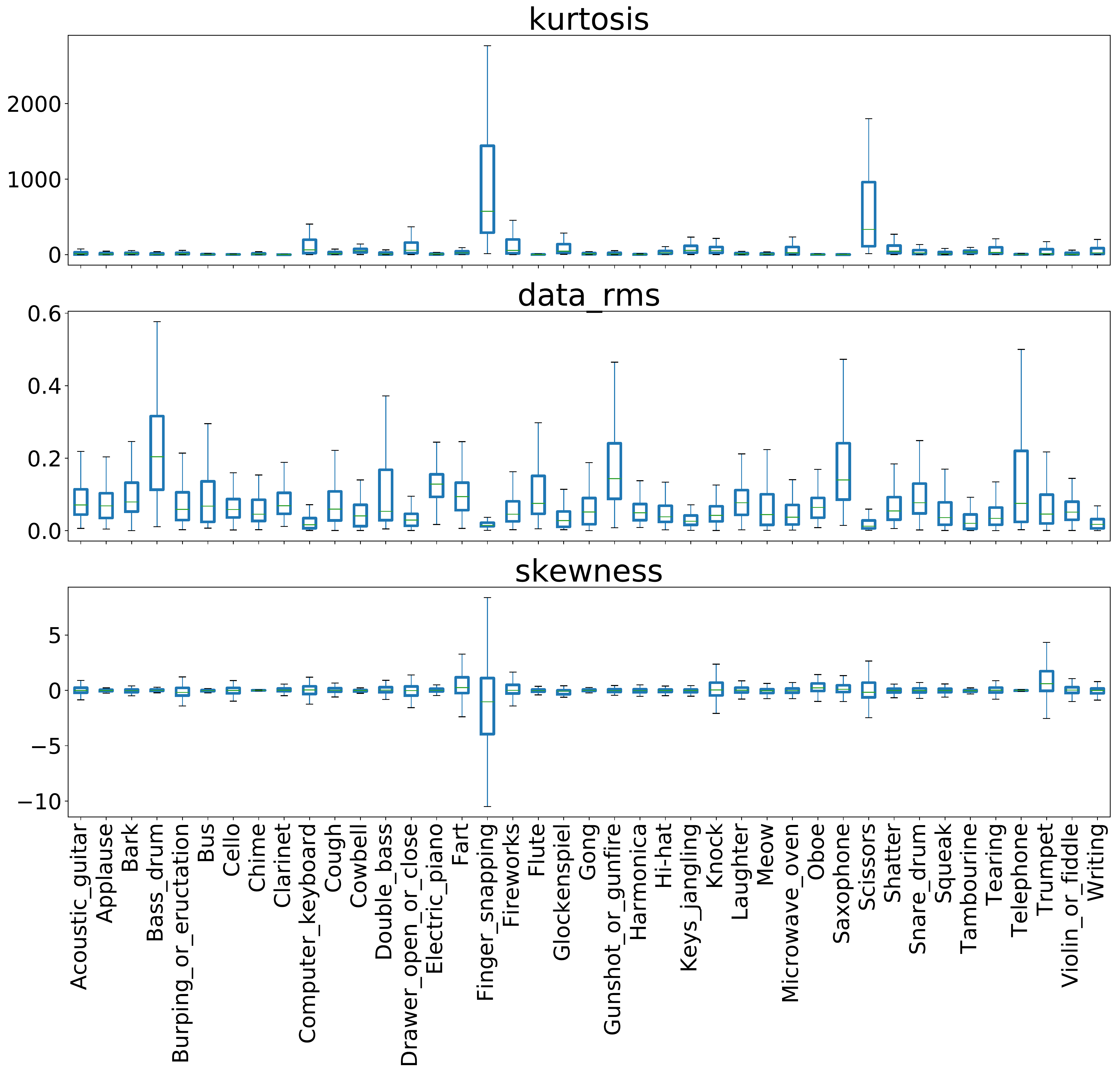}
	\caption{Sample statistical features for different categories (kurtosis, RMS and skewness values).}
	\label{Kurtosis}
\end{figure}

Sample statistical features are shown in Fig. \ref{Kurtosis}, in which the kurtosis, root mean square (RMS) and skewness of audio data are given. As can be seen from the figure, the kurtosis values varies with different categories, such as \emph{Finger\_snapping} and \emph{Scissors} have larger kurtosis values than other categories, \emph{Scissors} and \emph{Writing} have lower RMS values than other categories. The classifier could benefit from the combination of these statistical features. Thus, an effective approach to employ these patterns maybe can further boost the performance of the tagging task, which will be demonstrated in next section.

In this part, we explore to extract handcrafted statistical features, which will be deployed by the ensemble learning. Here, we are not aim to extract all of state-of-the-art handcrafted features for the classification task, but can be used to provide a demo for the ensemble learning framework.

\subsection{Ensemble learning}
Due to the limited size of DCASE 2018 Task 2, single model is easily overfitted. Ensemble different models can improve the accuracy and robustness for the classification task \cite{eghbal2016cp} using the complementary prediction result from different models. However, the ensemble learning has been under-explored for the audio tagging task. Most of previous methods simply average the predictions \cite{eghbal2016cp}. In this paper, we explore the use of stacked generalization in multiple levels to improve the accuracy and robustness to solve the audio tagging problem. The framework is computational, scalable and it have been tested on multiple machine learning tasks \cite{deng2012scalable}. Fig. 2 shows the proposed stacking architecture used in our task, which is composed of two levels. Level 1 consists of the deep models using different CNN architectures. Level 2 is shallow-architecture classier using the meta-features obtained from level 1.
Fig. 2 shows that both of deep learning-based meta features and handcrafted statistical features are used for the ensemble learning in level 2.

\begin{figure}[ht]
	\centering
	\includegraphics[width=0.48\textwidth]{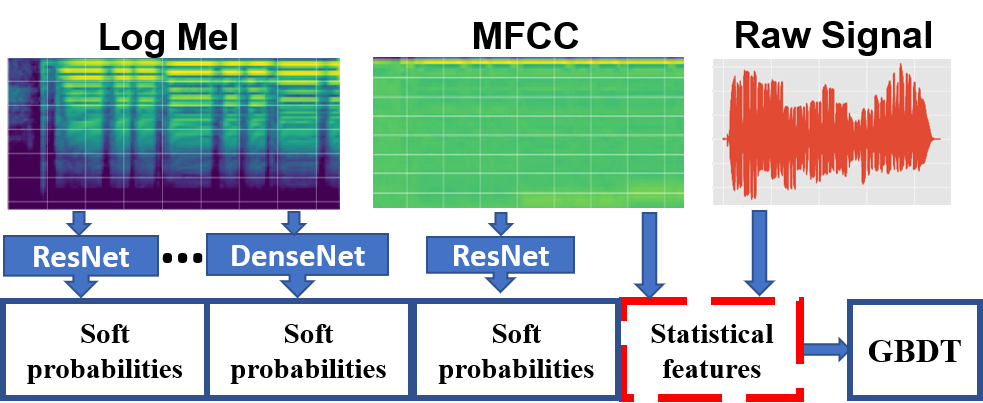}
	\caption{Framework of proposed ensemble learning approach.}
	\label{Framework}
\end{figure}

We randomly split the training data into 5 folds in our experiments. For the deep models, the out-of-fold based approaches are used to generate the out-of-predictions. All deep models use the same folds split configuration during the meta-feature creation. For each CNN, we run the CNN models for each out-of-fold training data, and one model to predict the probabilities for each sample in the validating set by using the whole training dataset. The predicted probabilities of different classes will be concatenated to generate meta-features. For each classifier, the probabilities for 41 classes will be used as the meta-features, which will be concatenated to generate the new training dataset (as can be seen in Fig. 2.), and the meta features will be used as the input for level 2.

For the ensemble learning in level 2, we employ the Gradient Boosting Decision Tree (GBDT) \cite{Friedman2001Greedy} for the task. The reason is that: compared to other approaches such as linear regression and support vector machine, GBDT provides better classification performance on several public machine learning challenges. GBDT is a tree-based gradient boosting algorithm. By continuously fitting the residuals of the training samples \cite{ke2017lightgbm}, each new tree reduces the errors produced from the prediction of the previous tree. The strategy of reducing residuals greatly improves the prediction accuracy of the model.

\subsection{Sample re-weight}
The objective function in ensemble learning is:
$Obj=\sum_{i=1}^{n}l(\hat{y_{i}(w)},y_{i})+\Omega(w)$,
where $l$ is the convex loss function, $\Omega$ is regulation component, including $L1$ regulation and $L2$ regulation. $n$ is the number of the samples, $\hat{y_{i}}$ is the prediction for sample $i$ and $y_{i}$ is the label. 
As the data size of the audio tagging is limited, some non-verified samples are employed as the training data, which may induce noisy samples. To train a classifier, the outliers in the training set have a high negative influence on the trained model. Indeed, designing supervised learning algorithms that can learn from data sets with noisy labels is an important problem, especially, when the data set is small. 

Here, we propose to induce a new hyper-parameter $r$ to re-weight the training samples. In more detail, the sample weight of manually verified samples is set as 1.0, while the weight for the non-manually verified samples are set as a constant value $r$ (smaller than 1). The best configuration for $r$ can be obtained using the grid search. Thus, the final objective function for ensemble learning can be re-written as:
\begin{equation}
Obj=\sum_{i=1}^{n_{\text{verified}}} l(\hat{y_{i}},y_{i})+\sum_{i=1}^{n_{\text{non- verified}}}r \times l(\hat{y_{i}},y_{i})+\Omega,
\end{equation}
where $r$ is the sample-wise weight, $n_{\text{verified}}$ is the number of manually verified audio clips and $n_{\text{non-verified}}$ is the number of non-verified audio clips.

\section{EXPERIMENTAL RESULTS}
\subsection{DATASETS}
The DCASE 2018 task 2 challenge dataset was provided by Freesound \cite{font2013freesound}. This dataset contains 18,873 audio files annotated with 41 classes of label from Google's AudioSet Ontology \cite{gemmeke2017audio}, in which 9,473 audio clips are used for training, 9,400 samples for validation (1,600 samples are manual verified).
The provided sound files are uncompressed PCM 16-bit, 44.1 kHz, mono audio files with widely varying recording quality and techniques. 
The duration of the audio samples range from 300 ms to 30 s due to the diversity of the sound categories.
The average length of the audio files is 6.7 seconds.
In the training dataset, the number of audio clips ranges from 94 to 300 depending on different classes.

\subsection{PREPROCESSING}

Two different kinds of inputs are employed to train the deep networks: log-scaled mel-spectrograms (log-mel) and Mel-frequency cepstral coefficients (MFCC) of the audio segment. For the raw signal, 1.5s audio segments are randomly selected. For the log-mel, we choose the number of the mel filter banks as 64, with a frame width of 80 ms and the frame shift is 10 ms. This will result in 150 frames in an audio clip. Then the delta and delta-delta features of log-mel is calculated with a window size of 9. Finally, the original log-mel features are concatenated with delta and delta-delta features to form a $3 \times 64 \times 150$ dimension \cite{feng2018sample}. MFCC follows similar generation procedure, except for the size. To prevent over-fitting, we apply mixup-data augmentation \cite{zhang2017mixup} with an ratio of 0.2 \cite{xu2018mixup}.

\subsection{Quantitative comparison between different CNN architectures}
We apply mean average precision (mAP) as evaluation criterion, the mAP@3 performance of CNN models is shown in Fig. 1. All the 1600 manually-verified samples are used for the evaluation. Fig. 3 shows that: (1) Using the same architecture, log mel feature achieves better mAP@3 than MFCC using all CNN architectures; (2) Using pre-trained model, deeper CNN models such as ResNext improve the mAP@3 for the tagging task with the prior knowledge extracted from the visual data. Moreover, the combination of Log mel and deeper model provide superior performance. (3) With the network pre-trained with the computer vision data, the CNN models can provide better performance with comparison to the model trained from scratch. This indicates that the size of the audio dataset might not be sufficiently large enough to train deep models from scratch.

\begin{figure}[h]
	\centering
	\includegraphics[width=0.45\textwidth]{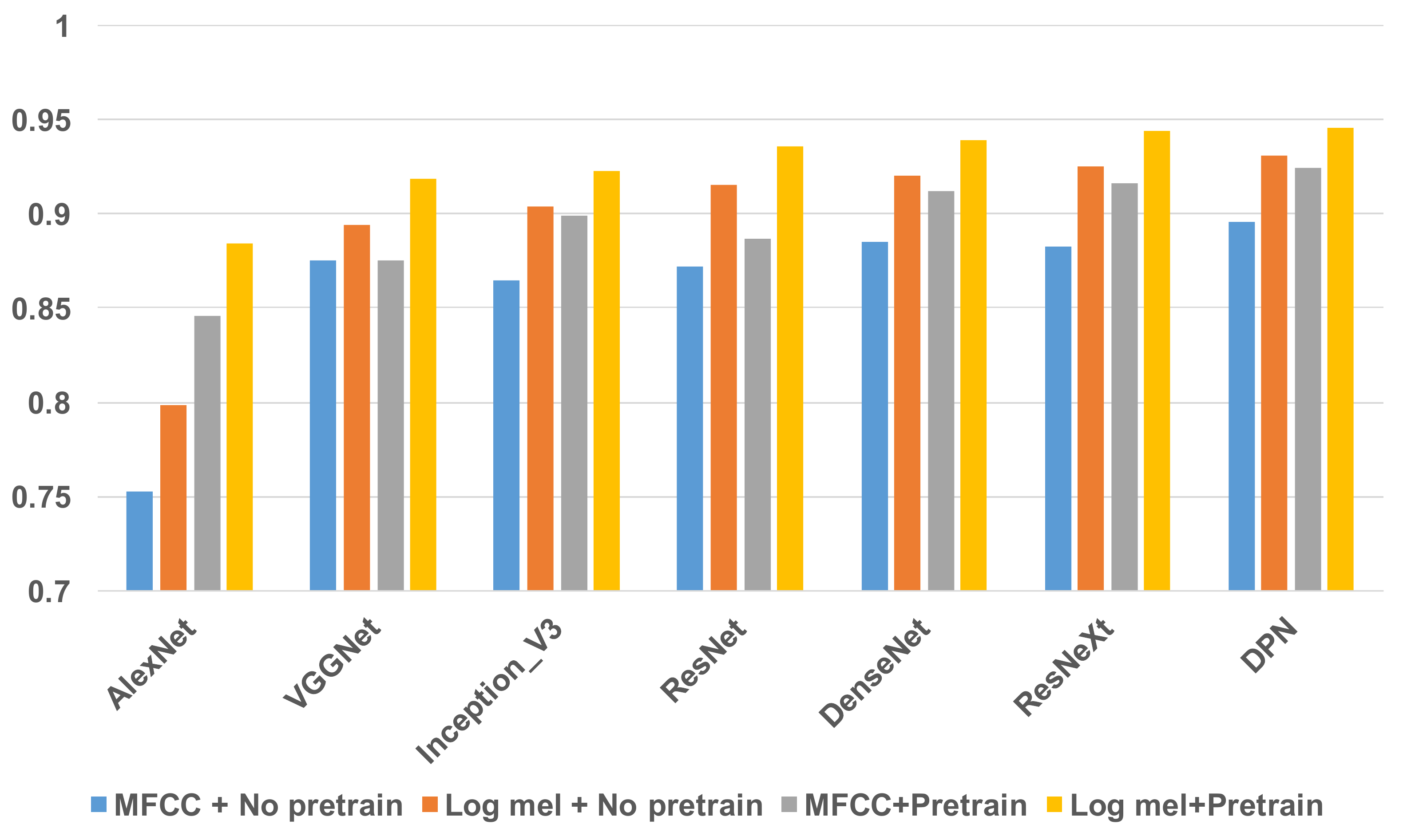}
	\caption{The mAP@3 obtained using different single CNN model.}
	\label{cnn_single}
\end{figure}

\subsection{Ablation study for statistical features}
To demonstrate the effectiveness of handcrafted features for proposed ensemble learning, we provide an ablation study for the handcrafted features. We first calculate the mAP@3 of ensemble learning by only using the out-of-fold predictions from deep models, which are regarded as a baseline. Thus, the handcrafted features are added to make a quantitative comparison with the same hyper-parameter configuration. As can be seen from Table 1, the obtained mAP@3s are much higher with statistical features.

\subsection{Ensemble learning with sample re-weight}
Out-of-fold predictions from the component models are aggregated to original file level before being fed into the level 2 model. Here, we implement our approach based on the LightGBM python library \cite{ke2017lightgbm}. 
The ‘max\_depth’ parameter of the model is set to 3 and learning rate was set at 0.03. which works good in our experiment. In addition, feature subsample and the sample subsample values were set at 0.7 to prevent from overfitting. Table 1 shows the experimental results with different $r$. As can be seen from the table, with handcrafted features, the mAP@3 of the classifier can be boosted with $r$ as 0.6.

\begin{table}
  \caption{The mAP@3 value of the tagging task different configuration for ensemble learning. (Statistical features is abbreviated as TF)}
  \vspace{6pt}
  \label{tab:result}
  \centering
  \begin{tabular}{p{1.5cm} p{1.5cm}p{2cm}}
    \toprule
		$r$& with TF & without TF\\
		\midrule
		0.0& 0.944& 0.935\\
		0.2& 0.946& 0.935\\
		0.4& 0.953& 0.946\\
		0.6& \textbf{0.958}& \textbf{0.947}\\
		0.8& 0.947& 0.943\\
		1.0& 0.946& 0.939\\
	\bottomrule
\end{tabular}
\end{table}

\section{Conclusion}
\label{sec:conclusion}
In this work, we proposed an novel ensemble-learning system employing a variety of CNNs and statistic features for the general-purpose audio tagging task in DCASE 2018. (1) A comparative study of the performance of different state-of-the-art CNN architectures are presented. (2) Statistical features are researched and are demonstrated to be effective for the ensemble learning. (3) The proposed ensemble-learning can employ the complementary information of deep-models and statistical features, which have a superior classification performance. (4) A sample re-weight strategy is employed to handle with the potential noisy label of the non-verified annotations in the dataset. Our system ranked the 1st and 4th out of 558 submissions in the public and private leaderboard of the DCASE 2018 Task 2 Challenge. For future work, we will evaluate the performance of our method on the Google AudioSet.

\section{Acknowledgment}
This work was supported by the National Grand R\&D Plan(Grant No. 2016YFB1000101). This study was also partially funded by the National Natural Science Foundation of China (No. 61806214). 

\bibliographystyle{IEEEtran}
\bibliography{refs}
\end{document}